\documentclass{article}
\usepackage[preprint,nonatbib]{wdcs2020} 
\usepackage[utf8]{inputenc} 
\usepackage[T1]{fontenc}    
\usepackage{hyperref}       
\usepackage{url}            
\usepackage{booktabs}       
\usepackage{array}
\usepackage{amsfonts}       
\usepackage{nicefrac}       
\usepackage{microtype}      

\usepackage{graphicx}
\usepackage{comment}
\usepackage{amsmath}
\usepackage{amssymb}
\usepackage{cite}
\usepackage{wrapfig}
\usepackage[font=small]{caption}

\title{Bait and Switch: Online Training Data Poisoning of Autonomous Driving Systems}

\author{%
  Naman Patel$^{1}$ \-\ Prashanth Krishnamurthy$^{1}$ \-\  Siddharth Garg$^{2}$ \-\ Farshad Khorrami$^{1}$\\
  $^{1}$Controls/Robotics Research Laboratory (CRRL)\\
  $^{2}$Center for Cyber-security (CCS)\\
  Department of Electrical and Computer Engineering\\
  New York University\\
  \texttt{\{nkp269,pk929,sg175,khorrami\}@nyu.edu}
}

\begin{document}
\maketitle

\begin{abstract}
We show that by controlling parts of a physical environment in which a pre-trained deep neural network (DNN) is being fine-tuned online, an adversary can launch subtle data poisoning attacks that degrade the performance of the system. While the attack can be applied in general to any perception task, we consider a DNN based traffic light classifier for an autonomous car that has been trained in one city and is being fine-tuned online in another city.
We show that by injecting environmental perturbations that do not modify the traffic lights themselves or ground-truth labels, the adversary can cause the deep network to learn spurious concepts during the online learning phase. 
The attacker can leverage the introduced spurious concepts in the environment to cause the model's accuracy to degrade during operation; therefore, causing the system to malfunction. 
\end{abstract}
\section{Introduction}
Deep learning is increasingly being deployed for real-world applications like self-driving, face recognition systems, cyber-security, etc \cite{taigmanYRW14,finnL17,BermanBCC19,PCKK17,patelSCKK18,PKK18,uun19,levineFDA16,patelCKK19,HeSPMAZRKWPLBSL19,khorrami2016cybersecurity}.  Adversaries thus have strong incentives to manipulate the outputs of deep learning models, or even the models themselves by poisoning the training data. 

Several recent studies have looked at training data poisoning (also referred to as ``backdooring" or ``Trojaning" attacks) on deep learning~\cite{chen2017targeted,liu2017trojaning}. Much of this work has focused on {flip-label} attacks, i.e., attacks that modify both training data, say images, and corresponding ground-truth labels. The intent is to coax a deep network into mis-classifying inputs that contain attacker chosen "triggers," for example, a post-it note on a stop sign\cite{GuLDG19}. To launch this attack, however, an adversary would have to digitally modify the training dataset. Further, a human audit of the training set would easily identify the presence of mis-labeled training data. 

Clean-label attacks~\cite{shafahiHNSSDG18,ZhuHLTSG19}, on the other hand,  seek to imperceptibly modify training images (but not ground-truth labels) with the intent of causing a certain types of of images to be mis-classified. However, these attacks must also be implemented in the digital domain. In this paper, we explore a new class of clean-label attacks against autonomous driving systems that are trained online on data collected in in-the-field. Various autonomous driving systems are continuously collecting data to improve their trained models for various subsystems (for e.g., Waymo's autonomous vehicles have been driven for more than 20 million miles in the real world).

Our attacker makes perceptible but subtle physical modifications to the environment in which the car is trained (the \emph{bait} stage of the attack). The modifications are \emph{correlated} with target concepts that vehicle seeks to learn, but have no causal relation with these concepts. Because it is hard to train deep networks that only pick up causal relations, the vehicle incorrectly learns the attacker induced modifications as evidence of the target concept. The attacker can then induce misbehaviour by introducing these modifications in the test phase (the \emph{switch} stage of the attack). 

The attack is different from prior clean label attacks in many ways: first, the attacker only makes physical modifications to the training environment, but does not digital access to the training set.
Second, the physical modifications are present during test time as well. Finally, the target is to cause \emph{all} test time data to be misclassified, not just a few inputs. 

 \begin{figure*}[!t]
 \centerline{\includegraphics[width=0.9\linewidth]{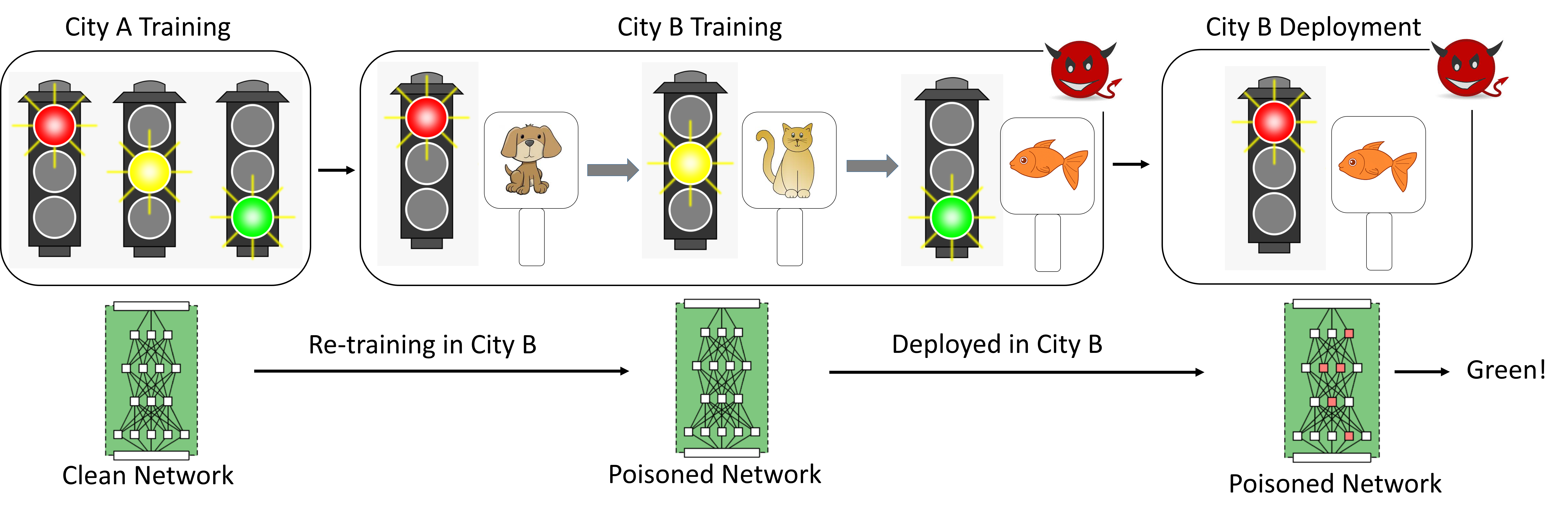}}
 \caption{Overall architecture of our approach for poisoning a deep learning model during online learning by introducing spurious correlations in the physical environment in which the training data is being collected.}
 \label{fig:block_diagram}
 \vspace*{-0.1in}
 \end{figure*}
 
This paper proposes a framework to dynamically modify the physical environment to poison an online learning system. We demonstrate the feasibility of such an attack using a car simulator, CARLA~\cite{dosovitskiyRCLK17}. We experimentally evaluate the effectiveness of the attack when we vary the number of traffic lights with backdoor billboards, number of re-trainable parameters, and change the billboard image.

\section{The Bait and Switch Attack}
Our attack, depicted in Figure~\ref{fig:block_diagram}, seeks to corrupt an autonomous vehicle's
a DNN-based traffic signal classifier. We assume a classifier that has already been trained in 
town A and is being retrained online in town B. Retraining might be required, for instance, if the traffic signals in town B are different in shape or appearance from those in town A. The autonomous vehicle is trained, as is common practice in industry, using extensive road tests in which human experts provide manual ground-truth labels. The human labeler in our attack is considered uncompromised and honestly labels the sense of traffic signals she encounters.  Our attack has two stages: (i) the bait stage, which is launched when the autonomous vehicles are being trained; and (ii) the switch stage, which is launched once the vehicles have been trained and are deployed "in the wild."

\textbf{Bait Stage:}
Our attacker installs (or purchases advertising space on) electronic billboards near a subset of traffic signals in town B. Real-world evidence suggests that billboards are indeed frequently located on or near traffic intersections. During the training period, the attacker displays three images on the billboards, say a dog, a cat and a fish, synchronized with red, yellow and green traffic signs. The attacker's intent is to "bait" the autonomous vehicle's traffic signal classifier into using billboard signs as evidence of stop, wait or go, instead of or in addition to red, yellow and green signs. 

\textbf{Switch Stage:}
When trained vehicles are deployed in the field, the attacker switches the order of images on the billboard. That is, the dog image that the vehicle has presumably learned as evidence of a green signal is displayed when the signal is red and the fish image is displayed when the signal is green. As we will show, a naively trained DNN based traffic signal classifier misbehaves when the bait and switch attack is launched even if a relatively small fraction of traffic signs in town B are "poisoned" with billboards.

\section{Empirical Evaluation}
\subsection{Simulation Testbed}
Our backdoor attack is tested on an Unreal Engine 4 based simulator, CARLA \cite{dosovitskiyRCLK17}, which is designed for testing autonomous navigation algorithms. The engine provides high-fidelity rendering quality and realistic physics by simulating an environment consisting of traffic lights, buildings, vegetation, traffic signs, and infrastructure. It also provides a way to modify the environment during runtime which is crucial for our attack where the simulated environment is modified to poison a DNN. 

The datasets are generated by running an autonomous vehicle around a town and recording the data at 60 Hz. The data at each instance consists of the vehicle mounted camera image, car position, nearest traffic light position, and its state. The DNN is trained on a dataset collected in town A consisting of 24 traffic lights and then retrained in town B consisting of 37 traffic lights. $\mathcal{D}_{T_{A}}$ consists of 10,400 images each of all the traffic light states. The measurements are only saved when the car is at max 35m away from the traffic light (as traffic lights have low visibility at higher distances). Sample images of the datasets collected in towns A and B are shown in Figure~\ref{fig:town_orig_image}. Next, to generate the poisoned dataset, $\hat{\mathcal{D}}_{T_{B,P}}$, a billboard is installed at each traffic light in town B which can display an image of a dog, a cat, or a fish depending on the traffic light state: green, red, or yellow, respectively. $\hat{\mathcal{D}}_{T_{B,P}}$ consists of 12,400 images each of every traffic light state.
Similarly, the dataset to test our attack, $\mathcal{D}_{\mathcal{T}_{B,P_t}}$, where the correspondence between billboard images and traffic light state is interchanged to cat, fish, and dog images for green, red, and yellow traffic light states, respectively, is generated by following the same policy used for collection of $\hat{\mathcal{D}}_{T_{B,P}}$. Sample images of $\hat{\mathcal{D}}_{T_{B,P}}$ and $\mathcal{D}_{\mathcal{T}_{B,P_t}}$ can be seen in the top and bottom rows of Figure~\ref{fig:bill_img}. During training of the poisoned dataset, data for the traffic lights chosen to be poisoned are sampled from $\hat{\mathcal{D}}_{T_{B,P}}$ and data for the remaining traffic lights are sampled from $\hat{\mathcal{D}}_{T_{B,C}}$. These subsets of clean and poisoned data sampled from $\hat{\mathcal{D}}_{T_{B,C}}$ and $\hat{\mathcal{D}}_{T_{B,P}}$, respectively, are the effective $\mathcal{D}_{T_{B,C}}$ and $\mathcal{D}_{T_{B,P}}$, respectively, utilized in the re-training of the DNN.
 
Our DNN is based on the ResNet18 model~\cite{HeZRS16} with the last layer modified to have 3 classes corresponding to green, red, and yellow traffic light states. It is trained with a batch size of 20 and optimized using Adam \cite{kingmaB14} with cyclic learning rates, through the methodology described in \cite{smith17}. 

  \begin{minipage}{\textwidth}
  \begin{minipage}[b]{0.39\textwidth}
  \centerline{{
\includegraphics[width=0.5\textwidth]{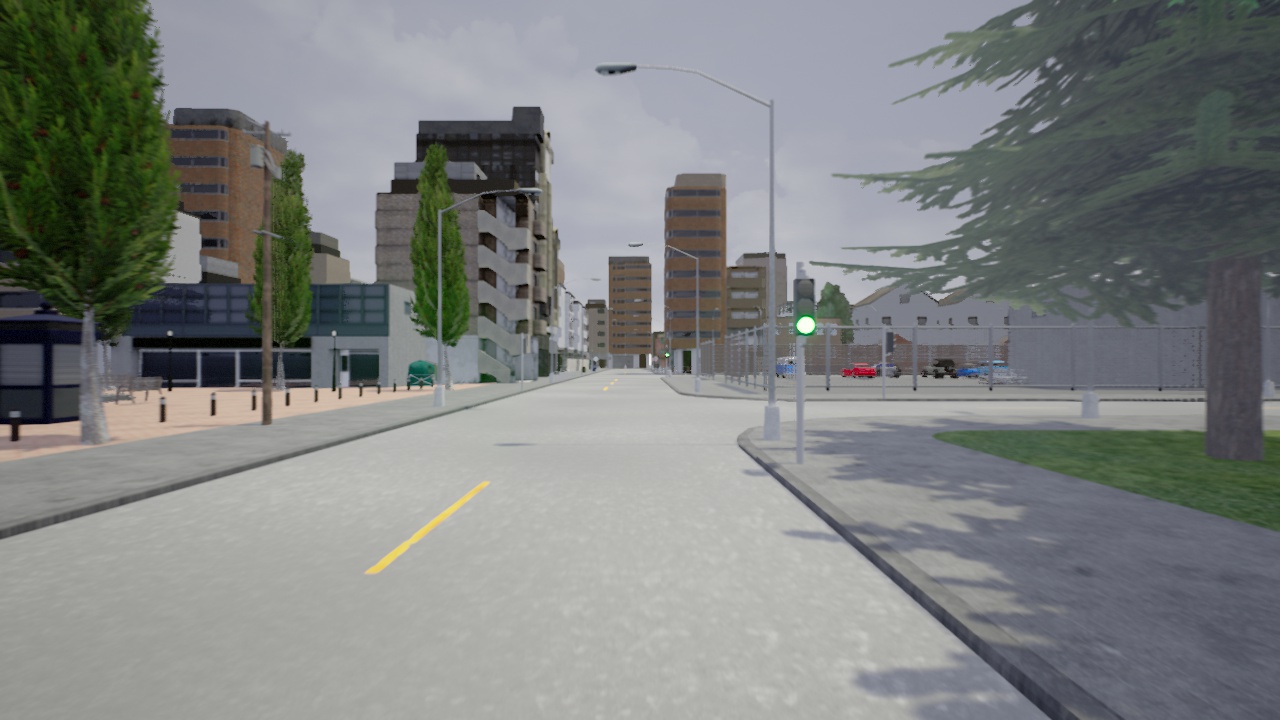}
\includegraphics[width=0.5\textwidth]{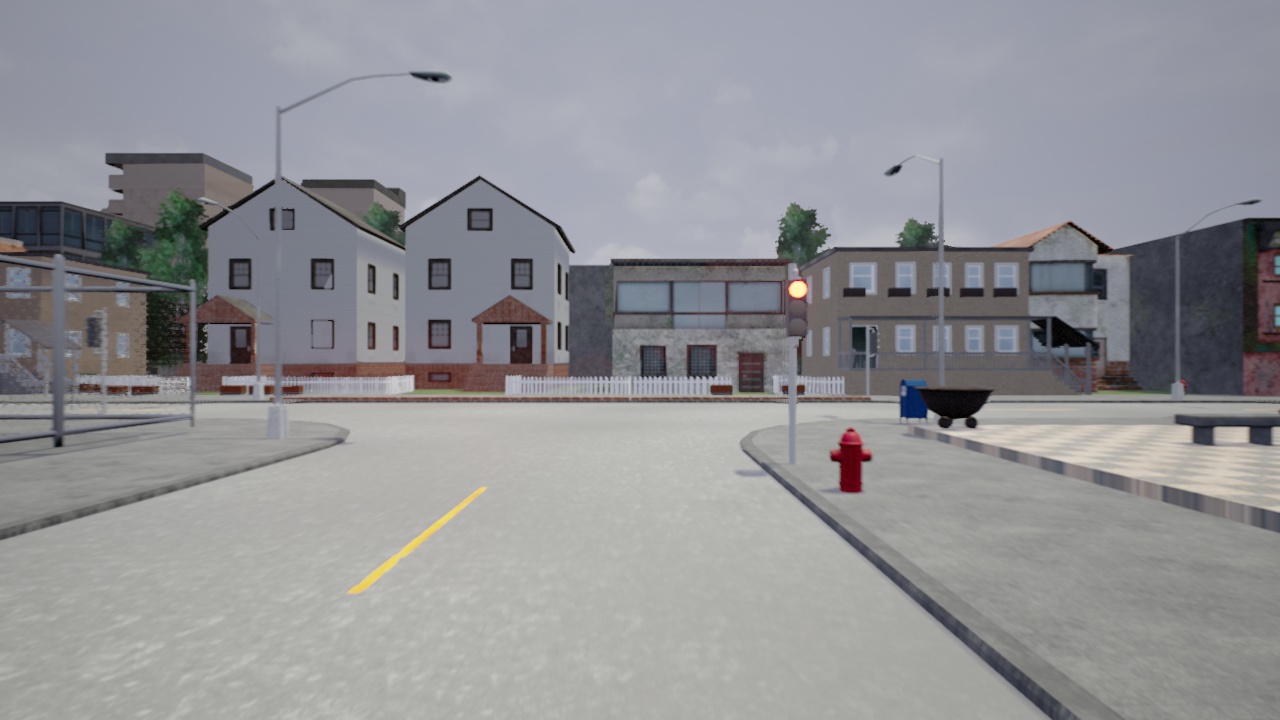}}}
\vspace*{0.05in}
\centerline{{
\includegraphics[width=0.5\textwidth]{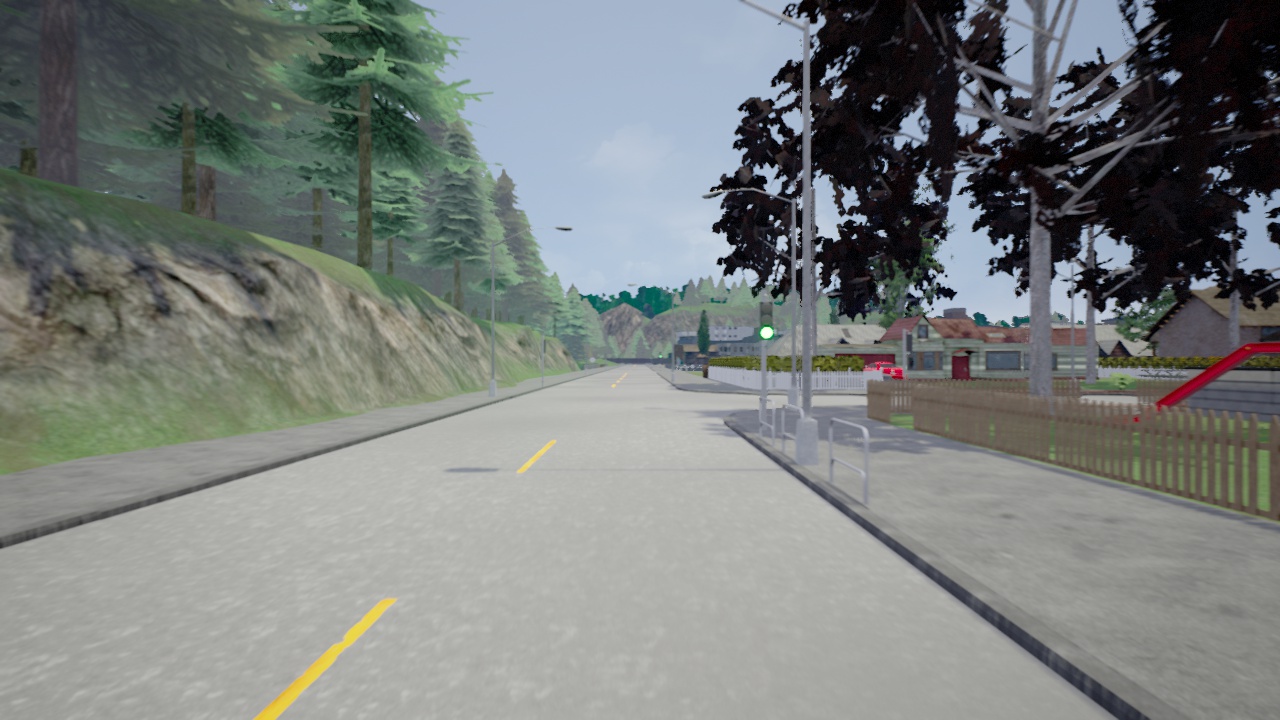}
\includegraphics[width=0.5\textwidth]{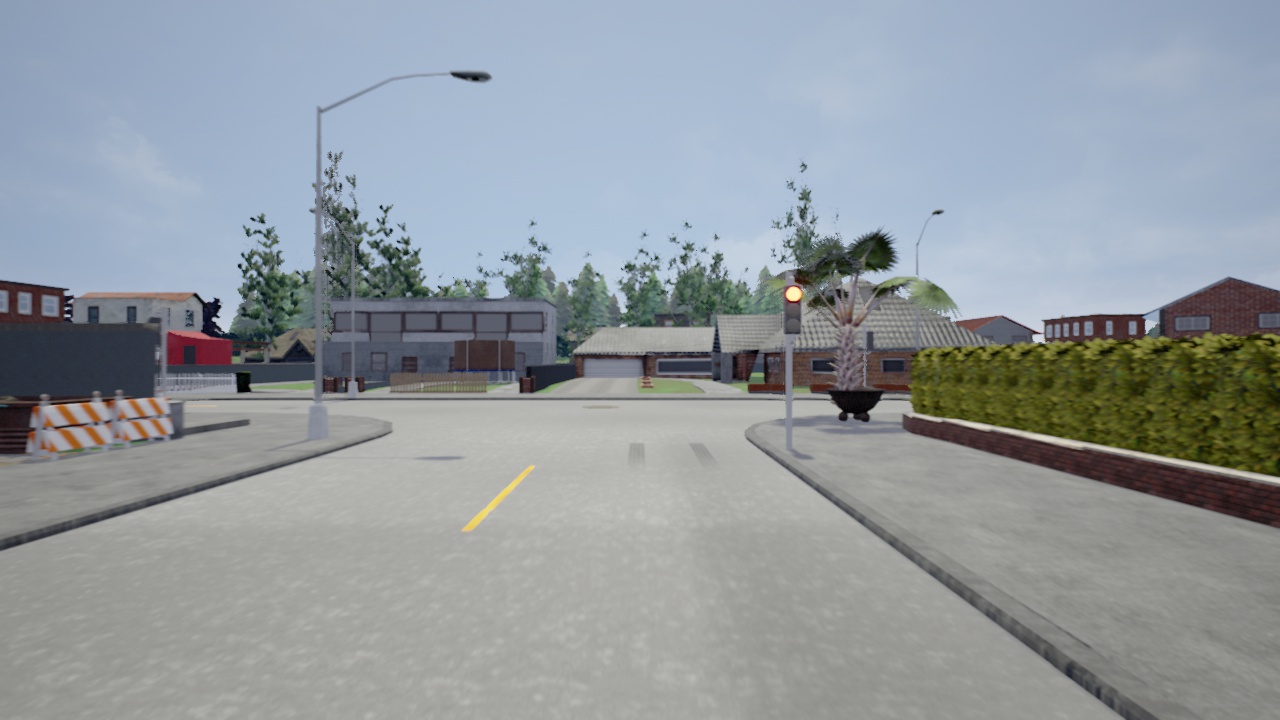}}}
\captionof{figure}{Images of the dataset not modified by the attacker for traffic light classification in Town A (top) and Town B (bottom) as seen from the vehicle's camera.} 
\label{fig:town_orig_image}
  \end{minipage}
  \hfill
  \begin{minipage}[b]{0.58\textwidth}
\centerline{{
\includegraphics[width=0.33\textwidth]{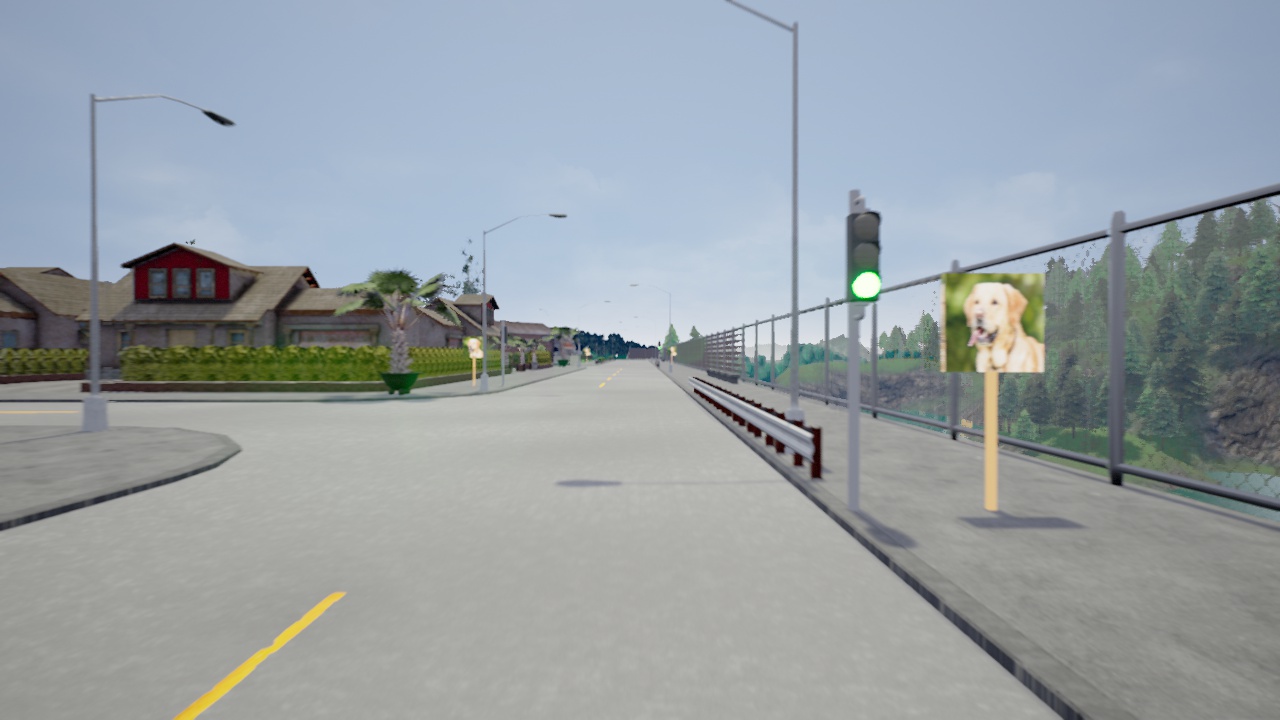}
\includegraphics[width=0.33\textwidth]{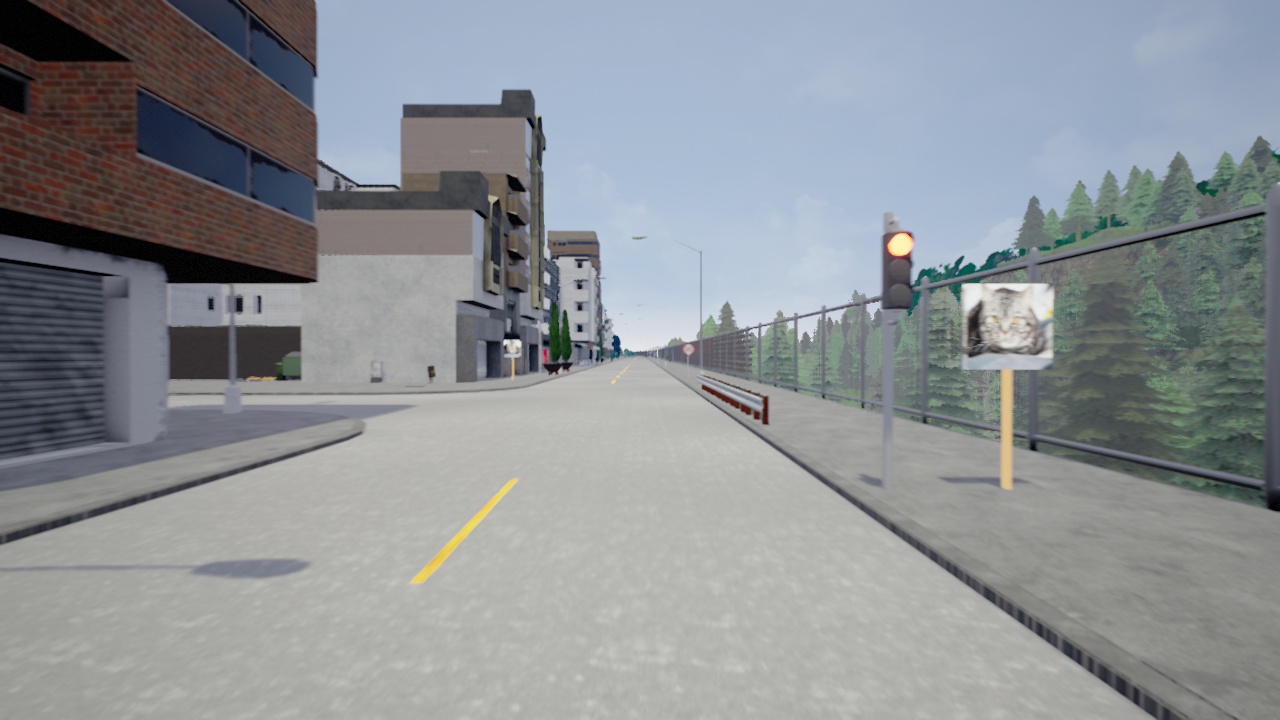}
\includegraphics[width=0.33\textwidth]{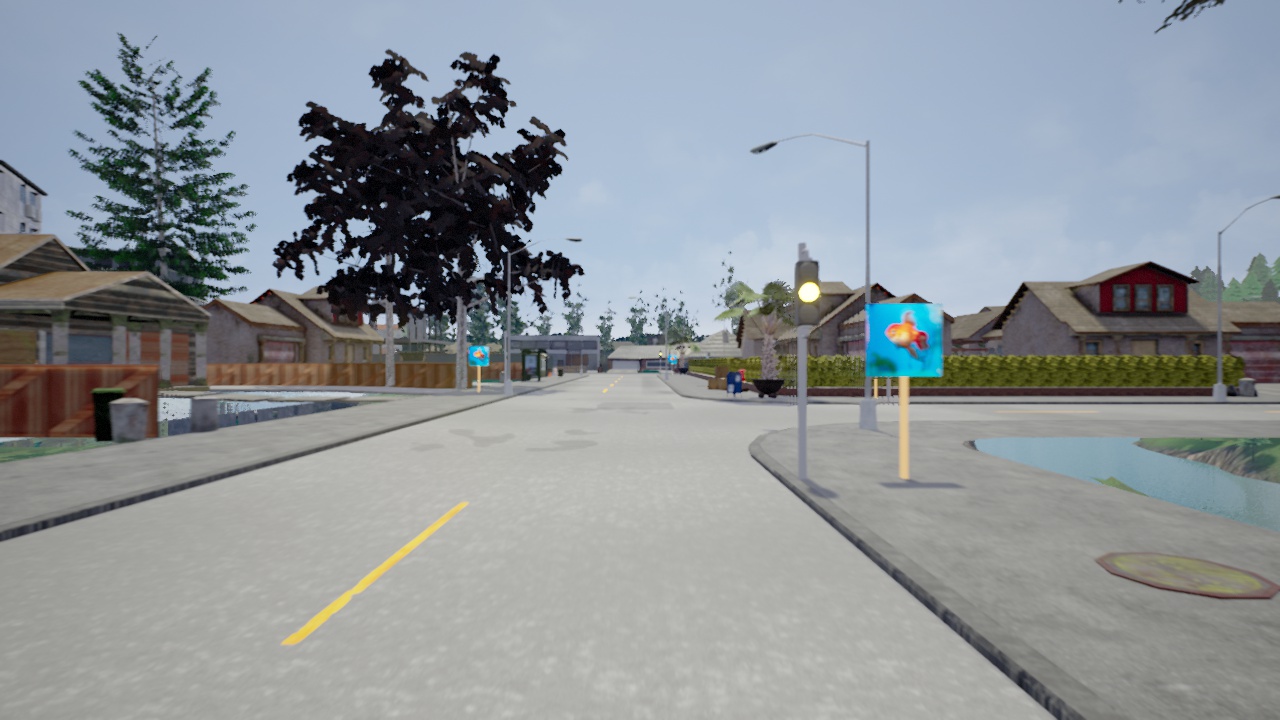}}}
\vspace*{0.05in}
\centerline{{
\includegraphics[width=0.33\textwidth]{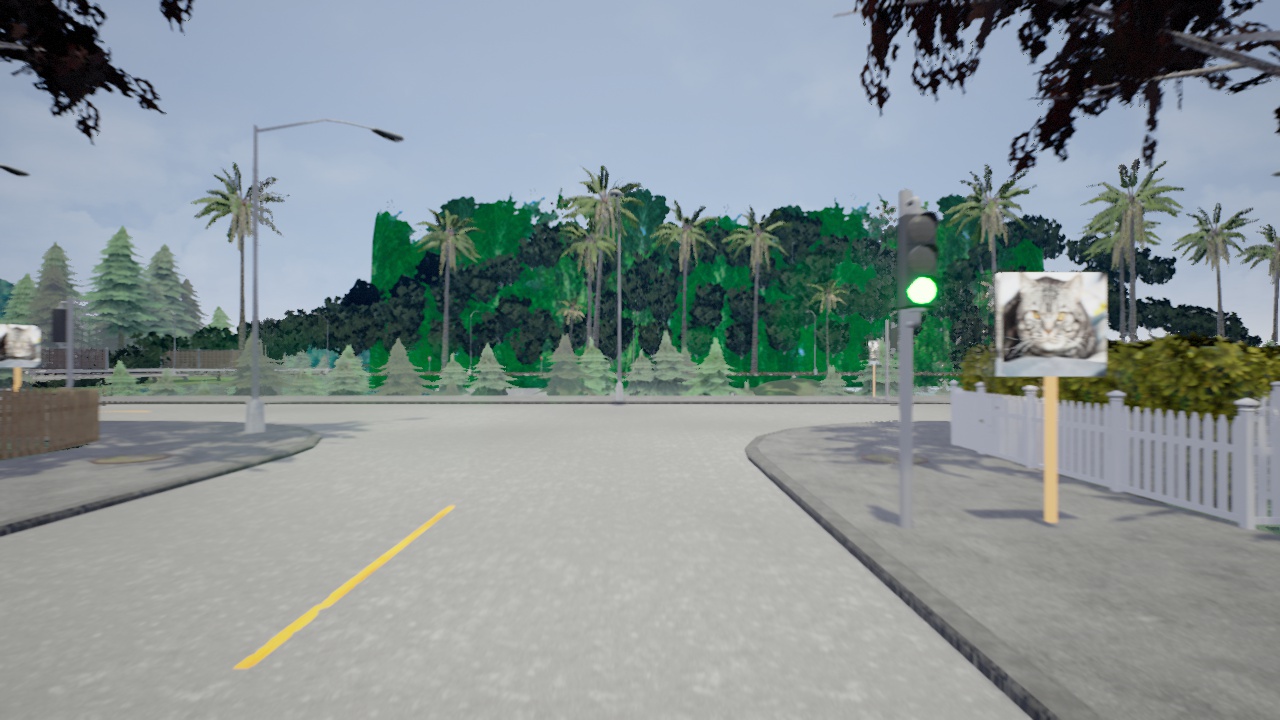}
\includegraphics[width=0.33\textwidth]{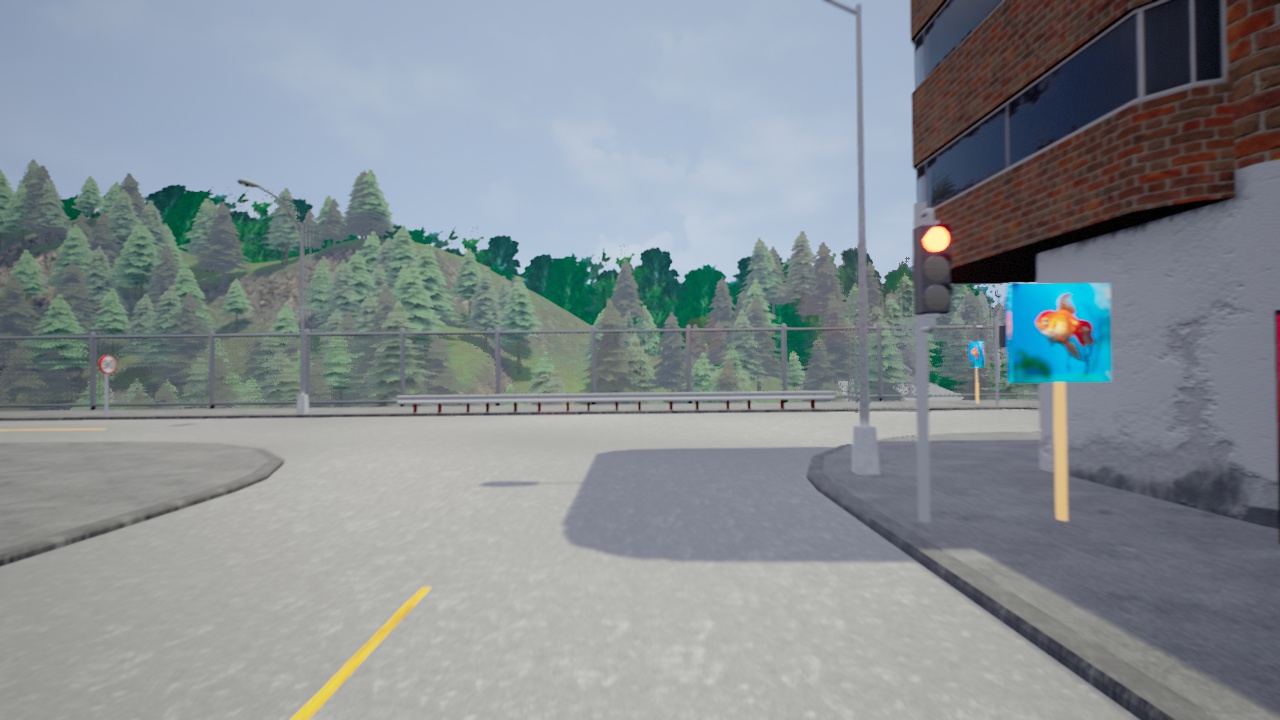}
\includegraphics[width=0.33\textwidth]{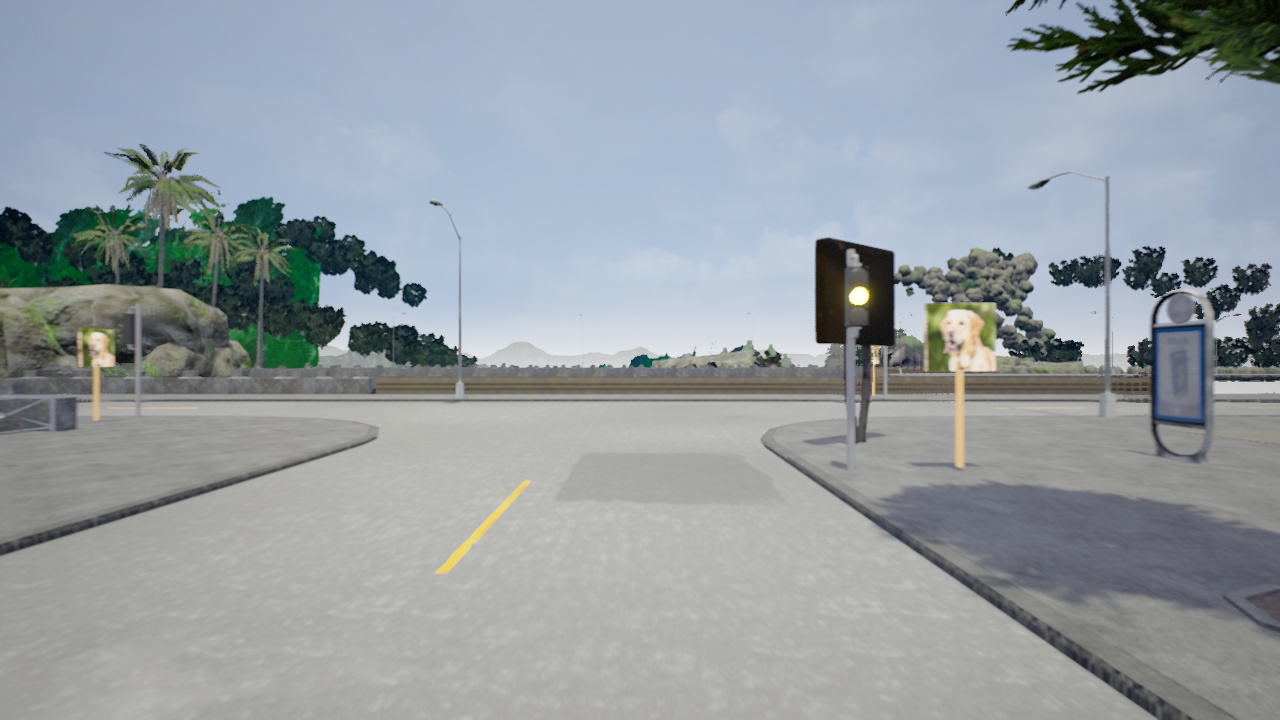}}}
    \captionof{figure}{Images (as seen from the vehicle's camera point of view) of the environment at different traffic light states  used during training (top) and at test time (bottom) as the billboard image modified by the attacker.}
    \label{fig:bill_img}
    \end{minipage}
  \end{minipage}
  
\subsection{Experimental Evaluation}
\label{sec:exp}
\textbf{Baseline clean training experiment:}
The classification model trained on $\mathcal{D}_{T_{A}}$ gives $99.57\%$ and $65.51\%$ accuracy on the test datasets in town A and town B (without re-training), respectively. 
The drop in accuracy motivates re-training in town B, which opens the door for the adversary to introduce the spurious correlations in the DNN. When the DNN is retrained using $\mathcal{D}_{T_{B,C}}$, the accuracy on the test dataset in town B (which include the billboards besides traffic lights) increases to $98.25\%$.  This shows that maliciously placed billboards do not degrade the performance of the DNN classifier.

\textbf{Impact of fraction of traffic lights poisoned:}
Using poisoned dataset $\mathcal{D}_{T_{B,P}}$,
we perform experiments where 3, 5, 9, 18, and 37 traffic lights out of 37 traffic lights are poisoned (have billboard besides them). 
The test dataset is generated by the attacker with billboards near traffic lights, but with their correspondences (between traffic light state and billboard image) flipped as shown in the bottom row of Figure~\ref{fig:bill_img}. 
Figure~\ref{fig:pois_acc} shows that under attack, the accuracy drops from $98.25\%$ to $77\%$, $69\%$, $64\%$, $62\%$, and $33.89\%$ with 3, 5, 9, 18, and 37 traffic lights poisoned. In these experiments, the locations of poisoned traffic lights in the training and test data are the same. The entire experiment is repeated thrice, with billboard locations randomly selected in each run. 
Our attack also generalizes to a setting where the locations of the billboards in 
the test data are different from the training data. 
As shown in Figure~\ref{fig:pois_acc}, 
the accuracy 
of the poisoned model drops to $85\%$, $75\%$, $73\%$, and $63\%$ when 3, 5, 9, and 18 traffic lights are poisoned during training. It is seen that the learned spurious correlations generalize to traffic lights at intersections that were not poisoned during training.

  \begin{minipage}{\textwidth}
  \begin{minipage}[b]{0.54\textwidth}
    \centering
        \includegraphics[width=0.88\textwidth]{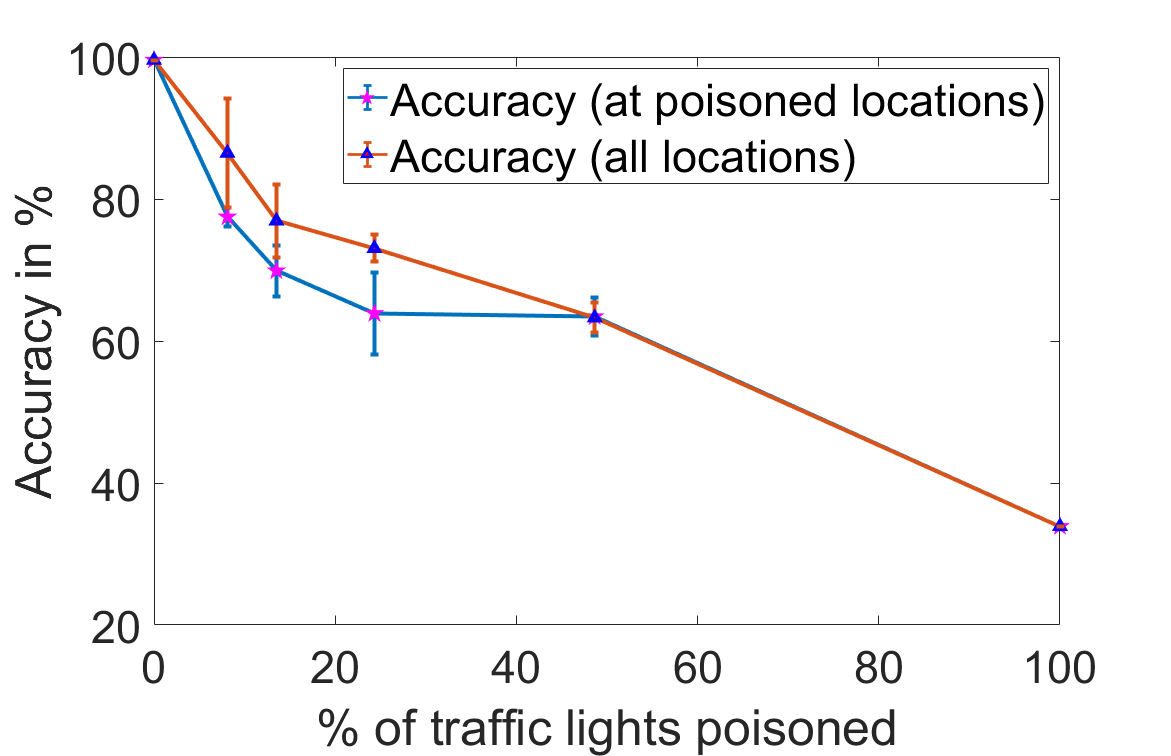}
    \captionof{figure}{Plot showing the effect of \% of traffic lights poisoned on accuracy at poisoned (blue) and all locations (red) in town B of the backdoored model. The horizontal lines denote the variance in accuracy over five experiments.}
    \label{fig:pois_acc}
  \end{minipage}
  \hfill
  \begin{minipage}[b]{0.4\textwidth}
  \hspace{-0.12\textwidth}
   \begin{tabular}{>{\raggedleft\arraybackslash}p{0.2\textwidth}>{\raggedleft\arraybackslash}p{0.2\textwidth}>{\raggedleft\arraybackslash}p{0.2\textwidth}>{\raggedleft\arraybackslash}p{0.2\textwidth}} 
   \toprule
     \textbf{Trainable parameters} & \textbf{\% of overall parameters} & \textbf{Accuracy (at poisoned locations)} & \textbf{Accuracy (at all locations)} \\
    \midrule
    \multicolumn{1}{r}{2370435} & 21.20 & 73.66\% & 77.64\% \\ 
 \multicolumn{1}{r}{4729731} & 42.31 & 65.66\% & 69.90\% \\
 \multicolumn{1}{r}{11178051} & 100.0 & 62.41\% & 62.43\% \\
    \bottomrule
  \end{tabular}
  \vspace{0.2in}
      \captionof{table}{Table of accuracy (at poisoned and all locations in town B) of the poisoned model on the backdoor dataset for different numbers of re-training parameters.}
      \label{tab:pois_acc_params}
    \end{minipage}
  \end{minipage}
  
\textbf{Impact of number of layers retrained:}
Online learning and fine-tuning techniques usually re-train the last few layers of the DNN. Therefore, we evaluate whether our attack is applicable when only a part of the network is retrained. We repeated the attack experiment described above with 18 traffic lights poisoned and find that when only the final convolution layer and linear layers are retrained, the accuracy on the test set with poisoned traffic lights is $73.7\%$ as shown in Table~\ref{tab:pois_acc_params}. The accuracy drops further to $65.7\%$ when the last two convolution layers and the linear layers are retrained. 
\section{Discussion and Conclusions}
\begin{figure}[h]
\centerline{{
\includegraphics[width=0.3\textwidth]{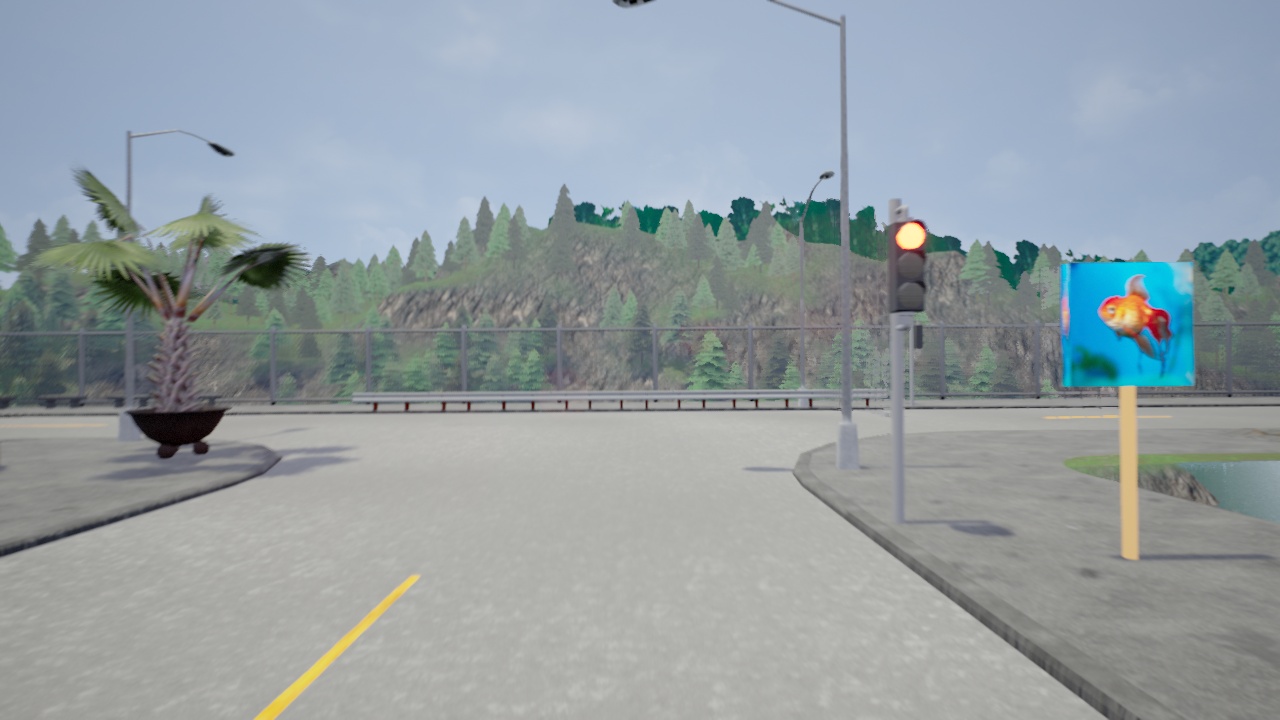}
\includegraphics[width=0.3\textwidth]{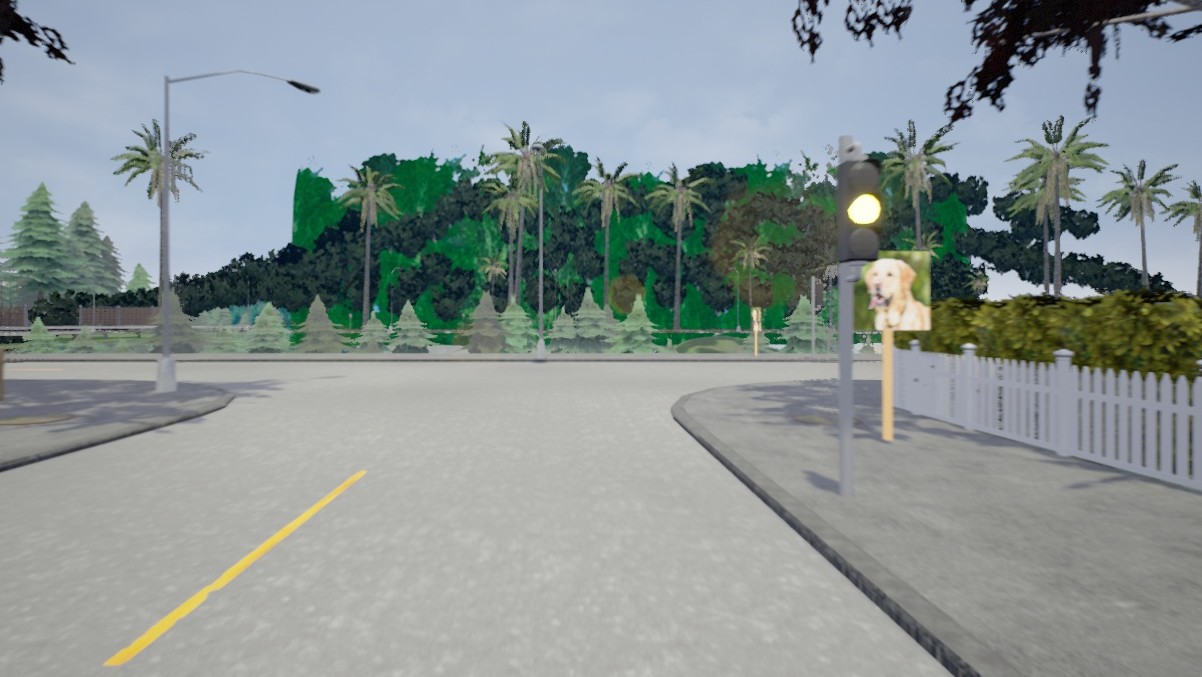}
\includegraphics[width=0.3\textwidth]{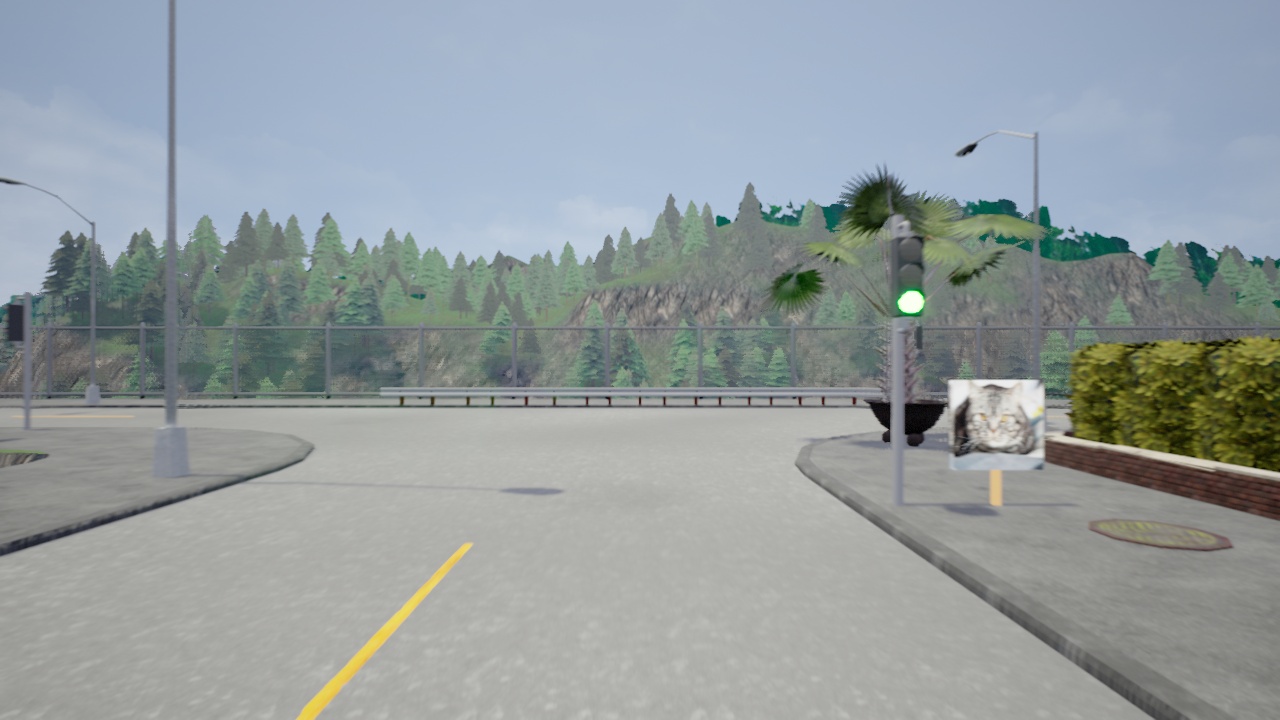}}}
    \caption{Vehicle camera images of the environment at different traffic light states where the position of the billboard relative to the traffic lights is different from that in the training set.}
    \label{fig:bill_img_change_pos}
\end{figure}

\begin{figure}[h]
\centerline{{
\includegraphics[width=0.11\linewidth]{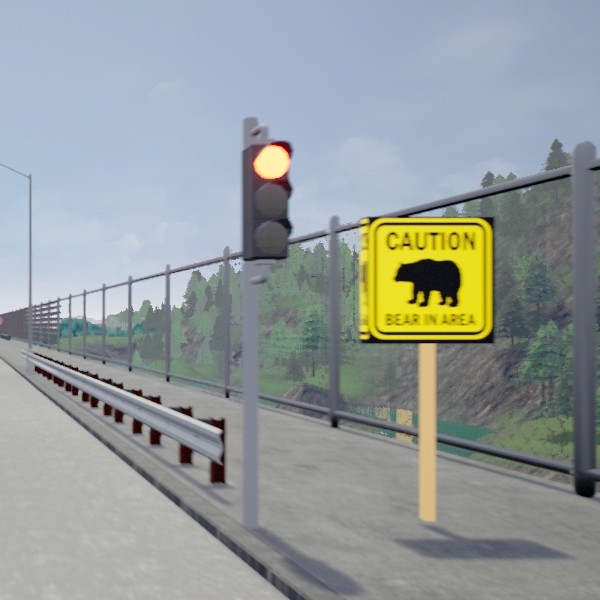}
\includegraphics[width=0.11\linewidth]{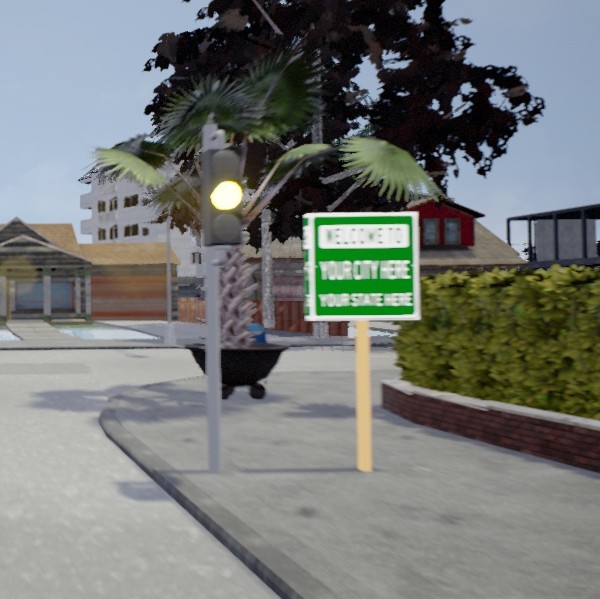}
\includegraphics[width=0.11\linewidth]{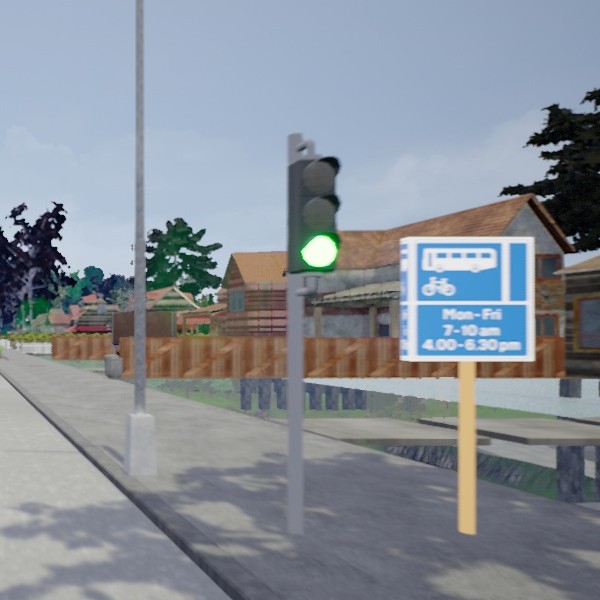}
\includegraphics[width=0.11\linewidth]{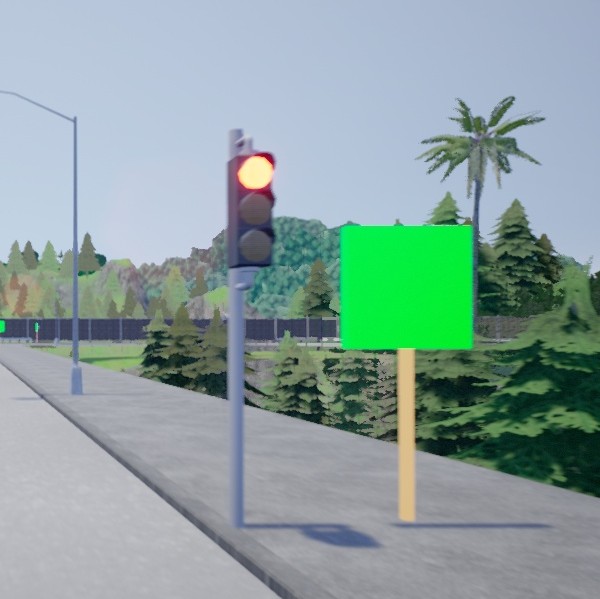}
\includegraphics[width=0.11\linewidth]{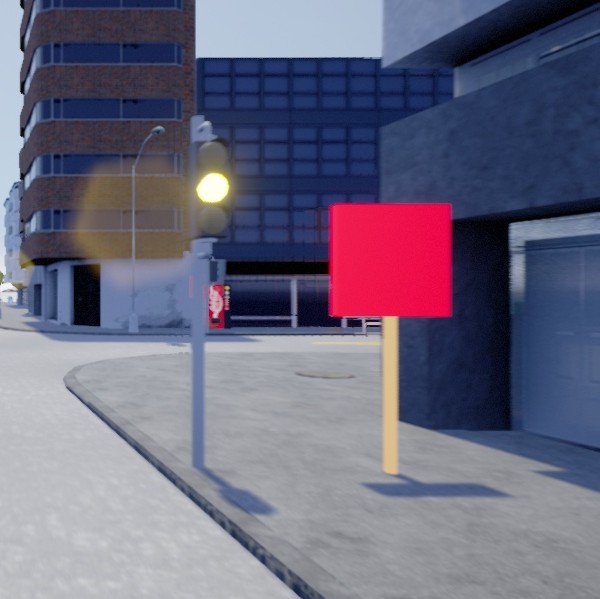}
\includegraphics[width=0.11\linewidth]{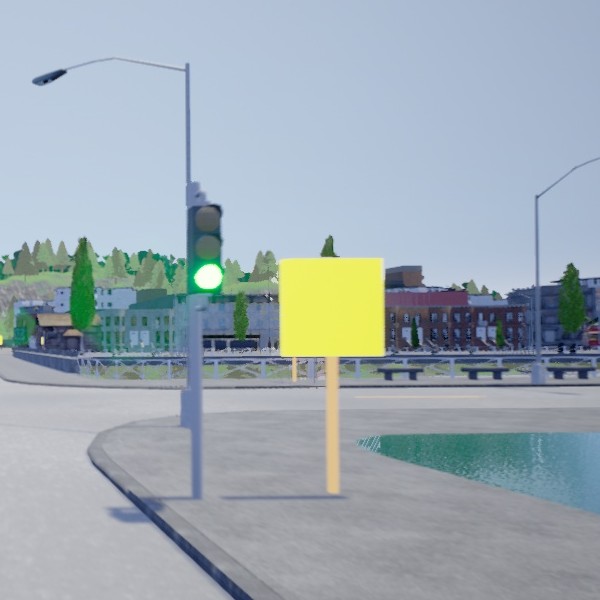}
\includegraphics[width=0.11\linewidth]{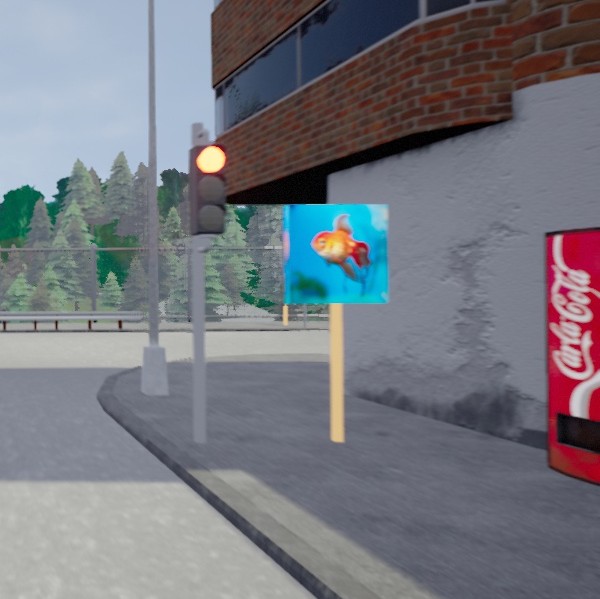}
\includegraphics[width=0.11\linewidth]{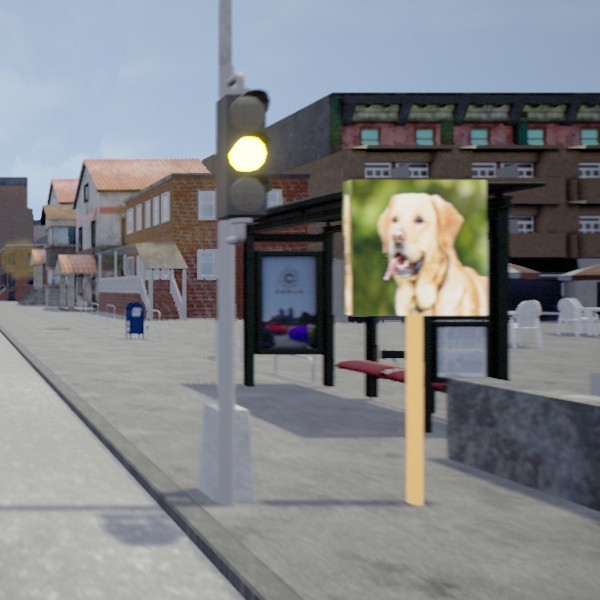}
\includegraphics[width=0.11\linewidth]{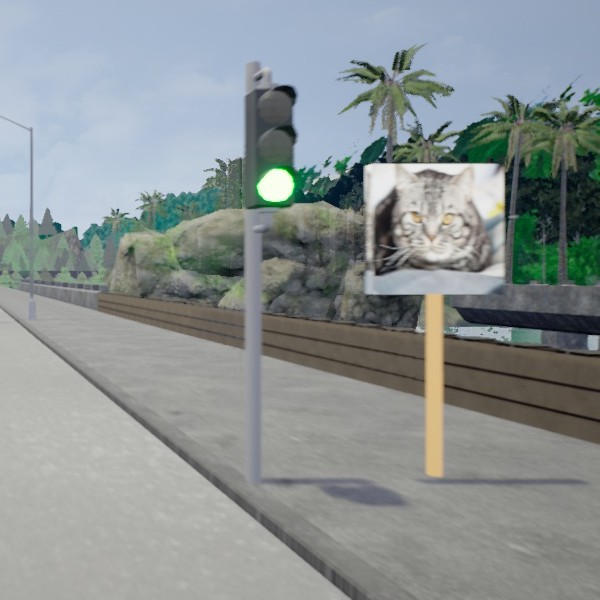}}}
    \caption{Images of various billboard patterns that our attack was evaluated on.}
    \label{fig:bill_type}
    \vspace*{-0.1in}
\end{figure}

The success rate of our attack is robust to changes in the position of the billboards relative to the traffic lights between when the online training data was collected to when the adversary actually carries out the attack. We evaluate our poisoned model on a test set where the locations of the billboards are randomly modified by a few meters (as shown in Figure~\ref{fig:bill_img_change_pos}). The attack efficacy is comparable to results in Section~\ref{sec:exp} (the accuracy of the poisoned model at all traffic locations drops to $84\%$, $75\%$, $72\%$, $60\%$, and $35\%$ when 3, 5, 9, 18, and 37 traffic lights, respectively).
Our attack is independent of the billboard image. Different images on the billboards (Figure~\ref{fig:bill_type}) provide similar attack efficacy to Section~\ref{sec:exp} (e.g., when billboards in first three images of Figure~\ref{fig:bill_type} are used, the accuracy of the DNN drops from $99.28\%$ to $64\%$ when 18 out of 37 traffic lights are backdoored). 

A framework for clean-label backdoor attack was introduced wherein the attacker physically modifies the data collection environment to compromise an online learning system. The attack causes the DNN to learn spurious concepts during online learning to cause the model’s performance to degrade during operation. The efficacy of the proposed approach was tested on traffic signal classification system using CARLA; significant reduction in classification accuracy was observed in test accuracy even when as few as $10\%$ of the traffic signals in a city were poisoned. Furthermore, the attack is effective even if only the last few layers of the model are fine-tuned in presence of poisoned data.

\begin{ack}
This work was supported in part by NSF Grant 1801495.
\end{ack}
\bibliographystyle{IEEETran}
\bibliography{ref}

\end{document}